\documentclass[pdflatex,sn-mathphys-num]{sn-jnl}

\usepackage{graphicx}%
\usepackage{multirow}%
\usepackage{amsmath,amssymb,amsfonts}%
\usepackage{amsthm}%
\usepackage{mathrsfs}%
\usepackage[title]{appendix}%
\usepackage{xcolor}%
\usepackage{textcomp}%
\usepackage{manyfoot}%
\usepackage{booktabs}%
\usepackage{algorithm}%
\usepackage{algorithmicx}%
\usepackage{algpseudocode}%
\usepackage{listings}%

\usepackage{dsfont}
\usepackage{amsmath}
\usepackage{booktabs} 
\usepackage{amsthm}
\usepackage{multirow} 
\usepackage{algorithm}
\usepackage{algorithm}
\usepackage{algpseudocode}
\usepackage{amsmath}
\usepackage{amssymb}
\algtext*{EndFor}
\algtext*{EndIf}
\usepackage[table]{xcolor}
\usepackage{booktabs}
\usepackage{array}
\usepackage{booktabs}

\usepackage{placeins}

\theoremstyle{thmstyleone}%

\theoremstyle{thmstyletwo}%

\theoremstyle{thmstylethree}%
\newtheorem{definition}{Definition}%

\begin{document}

\title[Article Title]{Path Planning with Motion Primitives in Dynamic Environments: SIPP on Lattices}

\author[1]{\fnm{Marat} \sur{Agranovskiy}}\email{agrinscience@gmail.com}

\affil[1]{\orgdiv{Markov Lab}, \orgname{Saint-Petersburg University}}

\abstract{Autonomous navigation in dynamic environments is a critical challenge, particularly when spaces are shared with other mobile agents whose future trajectories are known. While traditional grid-based planners efficiently find collision-free paths, their reliance on stop-and-turn mechanics over $2^k$-connected grids produces piecewise-linear trajectories that are kinodynamically highly sub-optimal for differentially constrained robots. In this paper, we present an adaptation of Safe Interval Path Planning (SIPP) that operates on state lattices, utilizing precomputed, kinodynamically smooth motion primitives. To efficiently handle dynamic environments, we rasterize the spatiotemporal swept volumes of moving obstacles directly onto the grid, treating grid cells as atomic units of space, whose resolution is typically dictated by inherent localization noise. We perform a comprehensive comparative analysis between our lattice-based approach and $2^k$-connected grid planners across diverse topological environments. Our evaluation considers a broad spectrum of performance metrics, including planning time, path angularity, cumulative heading change (angle-over-length), and bending energy. The results demonstrate that while the expanded state space of lattice-based search increases computational overhead, it yields trajectories with significantly superior kinodynamic properties. Specifically, our method achieves a reachability comparable to highly connected grids while ensuring smooth, continuous, and physically executable paths ready for real-world deployment.}


\keywords{Grid, A*, Path Planning, Safe Interval Path Planning, SIPP, State Lattice, Motion Primitives, Kinodynamic Constraints, Dynamic Environments\\
\\
\textbf{Code:} \url{https://github.com/PathPlanning/LatticeSIPP}\footnote{This is a preprint of a paper accepted for publication in the Proceedings of the\\International Conference on Interactive Collaborative Robotics (ICR 2026).}}

\maketitle

\section{Introduction}

Path planning in dynamic environments is a fundamental challenge in mobile robotics, particularly in shared workspaces like automated warehouses or multi-robot logistics hubs. In these settings, the ego-agent must often navigate around other autonomous robots whose future trajectories are deterministic and known a priori. While coordinating multiple agents to ensure global safety falls into the domain of Multi-Agent Path Finding (MAPF) \cite{ccbs} or Multi-Robot Motion Planning (MRMP) \cite{k_cbs}, treating these predictable agents as dynamic obstacles reduces the problem to a complex spatiotemporal search for a single agent. Even within this single-agent paradigm, computing safe, efficient, and physically executable paths remains computationally demanding.

Traditional grid-based planners typically rely on a discrete "stop-and-turn" control mechanism, yielding piecewise-linear paths that are geometrically valid but kinematically harsh for differentially constrained robots. To bridge the gap between abstract routing and real-world execution, some approaches apply post-processing techniques to convert jagged grid paths into feasible velocity schedules without altering the underlying geometry \cite{mapf_post}. While this ensures temporal feasibility -- such as allocating extra time for in-place rotations -- the spatial trajectory remains inherently angular. Conversely, a more robust methodology incorporates kinodynamic constraints directly into the search space. Works such as CL-MAPF \cite{cl_mapf} employ a Hybrid A* expansion to model Ackerman steering, while HCBS \cite{hcbs} accommodates mixed fleets of holonomic and non-holonomic agents.

A highly structured approach to embedding these kinodynamic constraints into the search is through state lattice planning \cite{pivtoraiko2005, pivtoraiko2009}. State lattices construct the search space using precomputed motion primitives: kinematically compliant continuous curves that seamlessly connect discrete robot states. However, the introduction of dynamic obstacles complicates this paradigm. Verifying continuous collisions between complex polynomial trajectories and moving constraints is analytically intractable in real-time. Consequently, contemporary algorithms often resort to temporal sampling, checking for collisions at discrete time intervals \cite{state_lattice_cbs, db_cbs}. This sampling strategy not only introduces significant computational overhead but also inherently risks missing inter-step collisions, thereby compromising safety guarantees.

To circumvent the pitfalls of temporal sampling, we leverage the inherent spatial discretization of practical robotic systems. In real-world deployments, grid resolution is not merely an algorithmic artifact but a fundamental operational limit dictated by sensor noise and localization uncertainty. Treating grid cells as indivisible, atomic units of space provides a natural and robust framework for collision detection. Inspired by the principles of MeshA* \cite{mesha_star}, which accelerates lattice search by reasoning over the shared swept cells of primitive bundles, we adopt a purely spatial-temporal rasterization approach. Instead of continuously evaluating or sampling polynomial curves, we rasterize the spatiotemporal swept volumes of both the moving obstacles and the ego-robot's motion primitives directly onto the grid.

Searching this augmented spatiotemporal domain ordinarily leads to a state-space explosion. To maintain computational tractability, we employ Safe Interval Path Planning (SIPP) \cite{sipp}. SIPP resolves the unbounded nature of the time dimension by aggregating contiguous, collision-free periods into discrete safe intervals, drastically reducing the branching factor. Its versatility has been proven across diverse applications, ranging from anytime \cite{anytime_sipp} and any-angle planning \cite{any_angle_sipp} to domains involving acceleration limits \cite{kinodynamic_sipp} and quadrotor navigation in uncertain environments \cite{quadrotor_sipp, safe_lattice_planner}.

In this paper, we present a synthesis of kinematically feasible state lattices and the SIPP algorithm to address dynamic path planning. We formulate a mathematically rigorous discrete state space and a sampling-free, primitive-based collision checking framework. To quantify the benefits of our approach, we empirically evaluate our lattice-based SIPP against straight-line planners operating on $2^k$-connected grids across various dimensionalities. Through a detailed analysis of search time, spatial routing efficiency, and critical geometric metrics -- specifically bending energy and trajectory angularity -- we demonstrate that an adequately expressive lattice control set can match the path lengths of highly connected grids while strictly guaranteeing physically executable, smooth trajectories.

\section{The Main Part}

\subsection{Problem Statement}
Consider a mobile agent with a physical footprint approximated by a circle of radius $R$, moving in a 2D workspace $W \subset \mathds{R}^2$. The workspace is tessellated into a regular grid with cells $(i,j) \in \mathds{Z}^2$.

\textbf{State Representation}. The state of the agent is defined by a 3D vector $(x, y, \phi)$, with coordinates $(x, y) \in W$ and heading angle $\phi \in [0, 360^\circ)$. We assume that the latter can be discretized into a set $\Theta = \{\phi_1, \dots, \phi_k\}$, allowing us to focus on discrete states $s = (i, j, \theta)$ that correspond to the centers of the grid cells $(i,j) \in \mathds{Z}^2$, with $\theta \in \Theta$.

\textbf{Environment Representation}. The environment contains both static and dynamic obstacles. To handle these efficiently, we utilize a hybrid paradigm:

\begin{enumerate}
    \item \textit{Static Environment}: Represented as a binary occupancy grid $S$, where each cell $(i,j)$ is either free space ($S_{ij}=0$) or a static obstacle ($S_{ij}=1$). We do not artificially dilate this grid by the agent's radius in advance to preserve the exact topological structure of the map (e.g., narrow corridors).
    
    \item \textit{Dynamic Environment}: Represented as a spatial-temporal map $D$. For each grid cell $(i, j)$, the map provides a set of disjoint, half-open time intervals $D_{ij} = \{[t_{in}^{(1)}, t_{out}^{(1)}), [t_{in}^{(2)}, t_{out}^{(2)}), \dots\}$ during which the cell is considered occupied by moving obstacles. To avoid computationally heavy continuous geometric intersections during the search, we pre-compute these intervals using the Minkowski sum. Specifically, the dynamic obstacles are geometrically dilated by the agent's radius $R$, swept along their continuous trajectories, and the resulting spatial-temporal swept volumes are rasterized onto the grid. Consequently, the circular agent is conceptually reduced to a point-mass: if the dilated moving obstacle covers the center of cell $(i,j)$ during a specific timeframe, it is recorded in $D_{ij}$.
\end{enumerate}

\textbf{Collision Checking Framework}. In practical robotic systems, localization and mapping sensors possess inherent noise. By choosing a grid resolution that aligns with this uncertainty margin, grid cells effectively become atomic units of space. Consequently, we cannot meaningfully distinguish between different sub-cell positions. Therefore, for the purpose of collision checking, we assume that whenever the agent occupies a cell, its geometric center is localized precisely at the cell's center. 

Under this assumption, static collision checking explicitly accounts for the agent's volume: a discrete state at $(i,j)$ is considered statically safe if all grid cells whose centers fall within the Euclidean distance $R$ from the center of $(i,j)$ are free in $S$. Any minor spatial discrepancies caused by the center-localization assumption are absorbed by a conservative padding of the agent's radius $R$. In contrast, because dynamic obstacles are already pre-dilated during the construction of $D$, dynamic collision checking is drastically simplified: we only need to query the safe intervals $D_{ij}$ at the single specific cell $(i,j)$ corresponding to the agent's center.

\textbf{Motion Primitives}. The kinematic constraints and physical capabilities of the mobile agent are encapsulated in \textit{motion primitives}, each representing a short, kinodynamically-feasible motion (i.e., a continuous state change).  We assume these primitives align with the discretization, meaning that transitions occur between two discrete states, such as $(i, j, \theta)$ and $(i', j', \theta')$. In other words, the motion starts at the center of a grid cell, ends at the center of another cell, and the agent’s heading at the endpoints belongs to a finite set of possible headings. Each primitive is additionally associated with a positive cost, which in our case is its execution time $\Delta t > 0$.

To evaluate dynamic collisions along a motion, each primitive defines a \textit{spatial-temporal collision trace}. For every cell $c=(i_c, j_c)$ traversed by the agent's center during the motion, the primitive dictates a relative time interval $[\tau_{in}^c, \tau_{out}^c) \subseteq [0, \Delta t)$ indicating when the agent's center resides within cell $c$. Note that $(i_c, j_c)$ are the coordinates relative to the primitive's start cell. 

For a given state $s=(i, j, \theta)$, there is a finite number of motion primitives that the agent can use to move to other states. Moreover, we assume there is at most one primitive connecting any specific pair of states, which corresponds to the intuitive understanding of a primitive as an elementary motion. We also consider that the space of discrete states, along with the primitives, is \textit{regular}; that is, for any $\delta_i, \delta_j \in \mathds{Z}$ if there exists a motion primitive connecting $(i, j, \theta)$ and $(i',j',\theta')$, then there also exists one from $(i+\delta_i,j+\delta_j,\theta)$ to $(i'+\delta_i,j'+\delta_j,\theta')$. While the endpoints
of such primitives are distinct, the motion itself is not. This
means that the collision traces of these primitives differ only
by a parallel shift of the cells, and their costs coincide.

Thus, we can consider a canonical set of \textit{template primitives}, the \textit{control set}, from which all other primitives can be obtained through parallel translation. We assume that such a set is finite and is computed offline. When a template primitive from the control set is instantiated at a discrete state $(i,j,\theta)$ at an absolute start time $t$, its relative temporal footprint is shifted to absolute time, yielding $[t + \tau_{in}^c, t + \tau_{out}^c)$ for each swept cell $(i+i_c, j+j_c)$. 

The execution of a primitive is \textit{collision-free} if all cells in its spatial trace are statically safe in $S$, and its absolute time intervals do not overlap with any dynamic obstacle intervals: $[t + \tau_{in}^c, t + \tau_{out}^c) \cap D_{i+i_c,\, j+j_c} = \emptyset$.

\textbf{Path and Objective}. A \textit{path} is a continuous sequence of collision-free motion primitives and, optionally, waiting actions (dwelling in a safe interval at the current cell). Adjacent primitives must share the same discrete spatial state to ensure continuity. The task is to find a collision-free path from a start state $s_0 = (i_0, j_0, \theta_0)$ at time $t=t_0$, to a goal state $s_f = (i_f, j_f, \theta_f)$. We wish to solve this problem optimally by finding the path that yields the minimum arrival time $t_f$ at the goal state.

\subsection{Safe Interval Path Planning with Motion Primitives}

To effectively navigate the spatial-temporal space defined by dynamic obstacles, we build upon the Safe Interval Path Planning (SIPP) algorithm \cite{sipp}. Standard search algorithms operating in the time-augmented state space suffer from a state space explosion due to the unbounded nature of the time dimension. SIPP mitigates this issue by abstracting the timeline into a strictly finite set of contiguous, collision-free periods called \textit{safe intervals}. We refer the reader to the foundational works \cite{sipp, anytime_sipp, any_angle_sipp} for a detailed explanation of the core algorithm as well as comprehensive examples, and now proceed with an overview of how SIPP is adapted to our framework with motion primitives.

\textbf{State Space}. Recall that the dynamic environment defines a set of occupied, half-open time intervals $D_{ij} = \{[t_{in}^{(1)}, t_{out}^{(1)}), \dots\}$ for each grid cell $(i, j)$. Consequently, the complementary free time naturally consists of a finite set of disjoint half-open intervals. Thus, for a cell $(i,j)$, a \textit{safe interval} $I = [t_{start}, t_{end})$ is defined as a maximal contiguous period of time during which the cell is not occupied by any moving obstacles from $D_{ij}$.

\begin{definition}[SIPP State]
A state in the SIPP search space is represented by a tuple:
\begin{equation*}
    p = \langle i, j, \theta, I \rangle, 
\end{equation*}
where $(i, j, \theta)$ is the discrete state and $I = [t_{start}, t_{end})$ is a safe interval at the cell $(i, j)$.
The minimum arrival time at this state is denoted as $t_{arr}$.
\end{definition}

Note that $t_{arr}$ serves as the $g$-value in the search graph (i.e., $g(p) = t_{arr}$). Because the agent is allowed to wait in place within a safe interval without colliding with dynamic obstacles, it does not have to depart exactly at its arrival time. Instead, the valid departure times from state $p$ form a continuous window $[t_{arr}, t_{end} - \varepsilon)$, where $\varepsilon$ represents the minimum time required for the agent to completely clear the current cell after initiating a motion primitive.

\begin{algorithm}[ht]
\caption{Constructing the Initial SIPP State}
\label{algorithm_initial_state}
\textbf{Input:} Initial discrete state $s_0 = (i_0, j_0, \theta_0)$, start time $t_0$\\
\textbf{Output:} Initial SIPP state $p_{start}$ or $\emptyset$\\
\textbf{Function name:} \textsc{GetInitialState}($s_0$, $t_0$)
\begin{algorithmic}[1]
\State $\texttt{SafeIntervals} \gets \mathds{R}_{+} \setminus D_{i_0 j_0}$ \Comment{Full planning horizon $\mathds{R}_{+} := [0, \infty)$}
\ForAll{$I = [t_{start}, t_{end}) \in \texttt{SafeIntervals}$}
    \If{$ t_{start} \leq t_0$ \textbf{and} $t_0 < t_{end}$}
        \State $p_{start} \gets \langle i_0, j_0, \theta_0, I \rangle$
        \State $g(p_{start}) \gets t_0$ \Comment{Initial arrival time is $t_0$}
        \State \Return $p_{start}$
    \EndIf
\EndFor
\State \Return $\emptyset$ \Comment{Start cell is occupied at $t_0$}
\end{algorithmic}
\end{algorithm}

\textbf{Initialization}. The search process begins at a user-defined start time $t_0$. To construct the root node of the search tree, we must map the initial configuration to a valid SIPP state. This involves querying the dynamic environment at the start cell to find the safe interval that encompasses $t_0$, as detailed in Algorithm \ref{algorithm_initial_state}. If the cell is occupied by a dynamic obstacle exactly at $t_0$, no such valid interval exists, and the problem is trivially unsolvable.

\textbf{Earliest Safe Departure}. The fundamental operation in our SIPP formulation is finding a valid, collision-free transition. Because motion primitives have a spatial-temporal footprint spanning multiple cells, verifying a transition requires more than a simple boundary check. We must find the \textit{earliest safe departure time} $t_{dep}$ within the departure window (possible departure times) such that the agent's footprint never overlaps with any dynamic obstacle.

Algorithm \ref{algorithm_earliest_departure} details this continuous time-search procedure. For each cell in the primitive's trace, we project the dynamic obstacles' occupied intervals back in time to compute \textit{invalid departure blocks}. By merging and sorting these blocks, we linearly scan for the first available departure time. Notably, this function depends solely on the template primitive and the grid cell (at which this primitive starts), decoupling it from the SIPP graph logic.

\begin{algorithm}[ht]
\caption{Continuous Earliest Safe Departure Search}
\label{algorithm_earliest_departure}
\textbf{Input:} Grid cell $(i, j)$, Primitive $\mathit{prim}$, Departure window $[t_{min}, t_{max})$\\
\textbf{Output:} Earliest valid departure time $t_{dep} \in [t_{min}, t_{max})$ or $\infty$\\
\textbf{Function name:} \textsc{GetEarliestDeparture}($(i,j)$, $\mathit{prim}$, $t_{min}$, $t_{max}$)
\begin{algorithmic}[1]
\State $\texttt{InvalidIntervals} \gets \emptyset$ \Comment{List of invalid departure bounds}
\ForAll{cell $c=(i_c, j_c)$ in $\mathit{prim}.\texttt{SpatialTrace}$}
    \State $[\tau_{in}^c, \tau_{out}^c) \gets \mathit{prim}.\textsc{GetRelativeTime}(c)$
    \ForAll{$[o_{start}, o_{end}) \in D_{i+i_c,\, j+j_c}$} \Comment{Occupied intervals of $c$}
        \State $i_{start} \gets o_{start} - \tau_{out}^c$ \Comment{Forbidden departure interval due to obstacle}
        \State $i_{end} \gets o_{end} - \tau_{in}^c$  
        \If{$i_{end} > t_{min}$ \textbf{and} $i_{start} < t_{max}$} \Comment{Keep only intervals that}
            \State $\texttt{InvalidIntervals}.\texttt{add}\left( [i_{start}, i_{end}) \right)$ \Comment{overlap departure window}
        \EndIf
    \EndFor
\EndFor
\If{$\texttt{InvalidIntervals}$ is empty}
    \State \Return $t_{min}$ \Comment{No obstacles threaten the execution}
\EndIf
\State \textbf{Sort} $\texttt{InvalidIntervals}$ in \textbf{ascending order by} $i_{start}$
\State $t_{dep} \gets t_{min}$
\ForAll{$[i_{start}, i_{end}) \in \texttt{InvalidIntervals}$}
    \If{$t_{dep} < i_{start}$}
        \State \Return $t_{dep}$ \Comment{Found a safe gap before the next obstacle}
    \EndIf
    \State $t_{dep} \gets \max(t_{dep}, i_{end})$ \Comment{Wait for the obstacle to pass}
    \If{$t_{dep} \ge t_{max}$}
        \State \Return $\infty$ \Comment{Departure window closed}
    \EndIf
\EndFor
\State \Return $t_{dep}$
\end{algorithmic}
\end{algorithm}

\textbf{Successor Generation}. During the A* expansion loop, generating successors relies on evaluating transitions into specific safe intervals of the target discrete states. For a given SIPP state $u$, we iterate over the canonical control set applicable to its orientation $\theta$. For each valid primitive, we determine the target discrete state and retrieve all its safe intervals. Crucially, we must attempt to find a valid departure time for \textit{each} individual target interval. For a specific target interval, we calculate a strictly bounded departure window $[t_{dep\_min}, t_{dep\_max})$ that satisfies both the temporal constraints of the current interval (clearing the starting cell before it closes) and the target interval (arriving while it is open). If the continuous departure search within this window yields a valid time, the successor state is constructed, as outlined in Algorithm \ref{algorithm_sipp_successors}.

\begin{algorithm}[ht]
\caption{Generating SIPP Successors}
\label{algorithm_sipp_successors}
\textbf{Input:} SIPP state $u$, arrival time $t_{arr}$ (current $g$-value)\\
\textbf{Output:} Set of successor tuples $(v, t_{arr}^{new}, \mathit{wait\_time})$\\
\textbf{Function name:} \textsc{SafeStatic}($(i,j), \mathit{prim}$)
\begin{algorithmic}[1]
    \ForAll{cell $c=(i_c, j_c)$ in $\mathit{prim}.\texttt{SpatialTrace}$}
        \State $(i_{abs}, j_{abs}) \gets (i+i_c, j+j_c)$  \Comment{Cell $c$ shifted by primitive's origin $(i,j)$}
        \If{$\exists (x,y) \text{ s.t. } \|(x,y) - (i_{abs}, j_{abs})\| \le R \text{ and } S_{x,y} = 1$}
            \State \Return $\texttt{False}$  \Comment{Obstacle within radius $R$}
        \EndIf
    \EndFor
    \State \Return $\texttt{True}$
\end{algorithmic}
\textbf{Function name:} \textsc{GetSuccessors}($u$, $t_{arr}$)
\begin{algorithmic}[1]
\State $i, j, \theta, I \gets u$
\State $\texttt{Successors} \gets \emptyset$
\ForAll{$\mathit{prim} \in \texttt{ControlSet}(\theta)$}
    \If{\textbf{not} $\textsc{SafeStatic}((i,j), \mathit{prim})$}  \Comment{Check instance of prim from (i,j)}
        \State \textbf{continue} \Comment{Skip if collision trace hits static obstacles}
    \EndIf
    \State $(i', j', \theta') \gets (i + \mathit{prim}.i_{goal},\, j + \mathit{prim}.j_{goal},\, \mathit{prim}.\theta_{goal})$ \Comment{Absolute target state}
    
    \State $\texttt{TargetIntervals} \gets \mathds{R}_{+} \setminus D_{i' j'}$
    \State $\tau_{out}^{start} \gets \mathit{prim}.\tau_{out}^{(0,0)}$ \Comment{Time to completely clear the starting cell}
    \ForAll{$I' \in \texttt{TargetIntervals}$}
        \State \Comment{Can't depart before current arrival, nor arrive before target interval opens}
        \State $t_{dep\_min} \gets \max(t_{arr}, I'.t_{start} - \mathit{prim}.\Delta t)$
        
        \State \Comment{Must clear current cell before $I$ closes, and arrive before $I'$ closes~~~~~~~~~~~~}
        \State $t_{dep\_max} \gets \min(I.t_{end} - \tau_{out}^{start}, I'.t_{end} - \mathit{prim}.\Delta t)$
        
        \If{$t_{dep\_min} \geq t_{dep\_max}$}
            \State \textbf{continue} \Comment{Invalid departure window for this target interval}
        \EndIf
        
        \State $t_{dep} \gets \textsc{GetEarliestDeparture}((i,j), \mathit{prim}, t_{dep\_min}, t_{dep\_max})$
        
        \If{$t_{dep} \ne \infty$}
            \State $t_{arr}^{new} \gets t_{dep} + \mathit{prim}.\Delta t$
            \State $\mathit{wait\_time} \gets t_{dep} - t_{arr}$
            \State $v \gets \langle i', j', \theta', I' \rangle$
            \State $\texttt{Successors}.\texttt{add}\left( (v, t_{arr}^{new}, \mathit{wait\_time}) \right)$
        \EndIf
    \EndFor
\EndFor
\State \Return $\texttt{Successors}$
\end{algorithmic}
\end{algorithm}

\begin{figure*}[th!]
\centering
\includegraphics[width=1\textwidth]{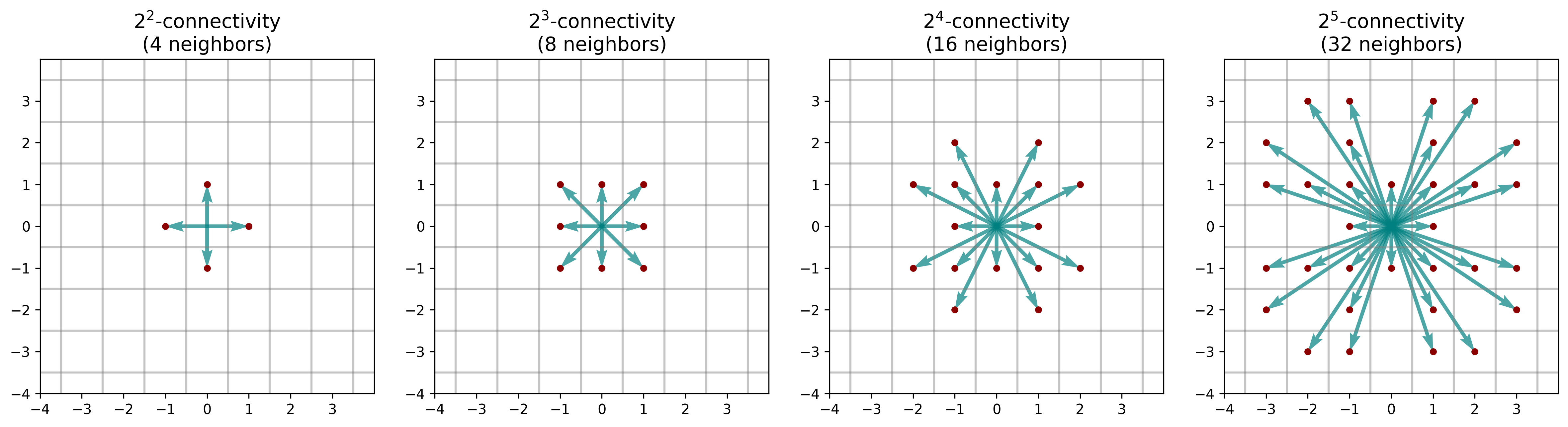}
\caption{Neighborhood topologies for $2^k$-connected grids ($k=2,3,4,5$).}
\label{2k_control_set}
\end{figure*}

\begin{figure*}[th!]
\centering
\includegraphics[width=1\textwidth]{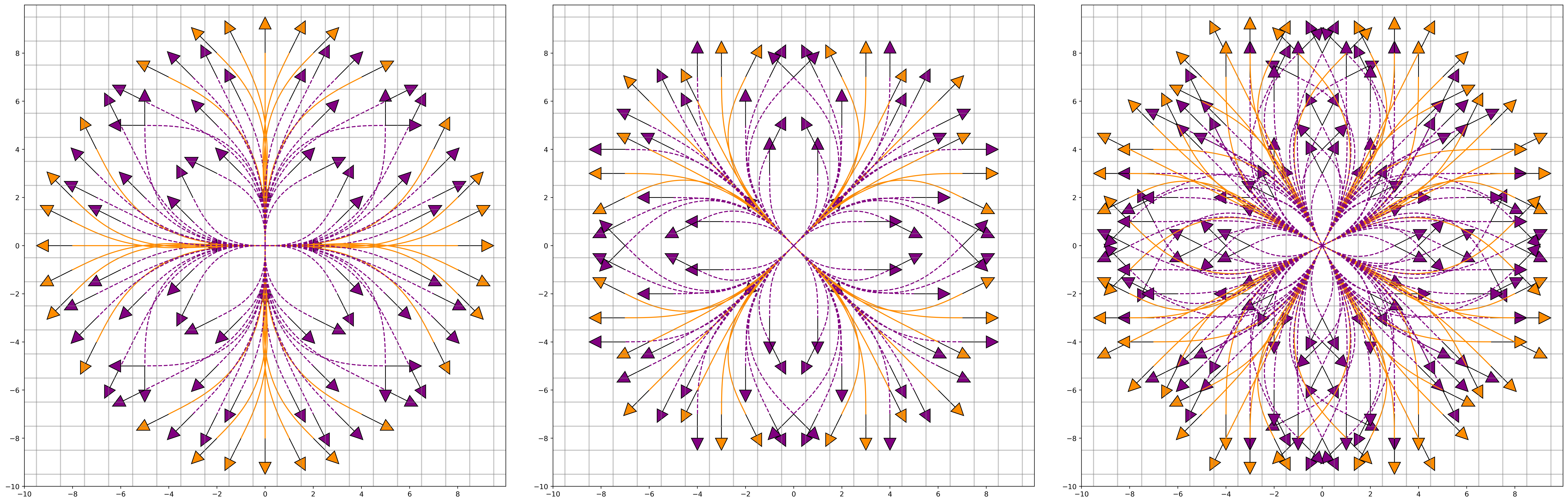}
\caption{Decomposition of the 16-heading control set for smooth transition motion planning. (Left) Cardinal and ordinal directions ($0^\circ, 90^\circ, \dots$); (Center) Diagonal primitives; (Right) Intermediate directions with $22.5^\circ$ angular resolution. Solid orange lines denote the \textbf{base primitive set} ($7 \times 16 = 112$ primitives), which is augmented by auxiliary primitives (dashed purple) to form the complete \textbf{extended control set} ($24 \times 16 = 384$ primitives).}
\label{smooth_control_set2}
\end{figure*}

\section{Results}

\subsection{Evaluated Algorithms}

To empirically evaluate the performance of SIPP across different topological structures, we utilize a simulated differential-drive robot. We benchmark the algorithm using six distinct motion primitive sets\footnote{Implementation and visualization source code: \url{https://github.com/PathPlanning/LatticeSIPP}.}:

\begin{itemize}
  \item \texttt{SIPP-Basic}: A minimal set of kinodynamically smooth motion primitives designed for seamless transitions between discrete states in a 3D space $(i, j, \theta)$ (see Fig.~\ref{smooth_control_set2}).
  \item \texttt{SIPP-Extended}: An enriched control set containing supplementary maneuvers to enhance maneuverability in constrained spaces (see Fig.~\ref{smooth_control_set2}, dashed lines).
  \item \texttt{SIPP-2\string^k ($k=2, 3, 4, 5$)}: A $2^k$-connected grid neighborhood where transitions are straight-line segments. Unlike the lattice sets, the state space is inherently 2D $(i, j)$, as the robot is assumed to be capable of moving to any valid neighbor regardless of its current orientation. The topological structures of the $2^k$ neighborhoods are illustrated in Fig.~\ref{2k_control_set}.
\end{itemize}

\subsection{Evaluation Metrics}

We compare the algorithms based on both search efficiency and the geometric quality of the resulting spatial trajectories. Our evaluation metrics are defined as follows:

\textbf{1. Search Time}\\
The computational time required by the A* expansion loop to find the optimal path.

\textbf{2. Path Cost}\\
The total execution time of the trajectory from the start state to the goal state. In the SIPP framework, this value naturally encapsulates both the traversal time along the primitives and the necessary waiting times (dwelling in safe intervals) to yield to dynamic obstacles.

\textbf{3. Trajectory Angularity ($\mathcal{A}$)}\\
To quantify the "jaggedness" of the spatial path, we sample points along the trajectory at a fixed high-frequency interval $\delta t$. For every consecutive triplet of sampled points $p_{i-1}, p_i, p_{i+1}$, we compute the discrete curvature, defined as the inverse radius of their circumscribed circle $R_c^{(i)}$. A larger inverse radius indicates a sharper, less natural turn, whereas collinear points yield an infinite radius and thus zero curvature. The total angularity is defined as the sum of these discrete curvatures along the trajectory:
$$\mathcal{A} =\sum_{i} \frac{1}{R_c^{(i)}}$$

\textbf{4. Angle-Over-Length (AOL)}\\
This metric quantifies the cumulative angular displacement \cite{benchMR-aol}. Consider a trajectory as a planar curve parameterized by its arc length $s \in [0, L]$. The infinitesimal change in the agent's heading $\theta$ at any point $s$ can be approximated as $|d\theta| \approx |\theta'(s)| ds$. In differential geometry, for a curve with a continuous tangent, the rate of change of the heading angle with respect to the arc length is exactly the signed curvature: $\theta'(s) = \kappa(s)$. Consequently, the total variation of the angle for smooth path with lattice primitives is the integral of the absolute curvature:

$$\text{AOL}_{smooth}\int_{0}^{L} |d\theta| = \int_{0}^{L} |\kappa(s)| ds$$

For the piecewise-linear paths produced by the $2^k$-connected grids, the heading derivative is zero along the segments but forms a Dirac delta function at the junctions. Thus, we compute this as the sum of the absolute angles between consecutive straight-line segments:
$$\text{AOL}_{grid} = \sum_{j} |\Delta \theta_j|$$

\textbf{5. Bending Energy ($E_{bend}$)}\\
The bending energy represents the physical effort required to execute the path, defined as the integral of the squared curvature $\kappa(s)$ along the arc length $s$:
$$E_{bend} = \int_{0}^{L} \kappa^2(s) ds$$
For the lattice sets, the primitives are designed to join with finite curvature, allowing this integral to be computed continuously. However, for $2^k$ grid paths, point-turns imply infinite instantaneous curvature. To provide a fair physical comparison, we model these junctions by approximating the point-turn as a continuous rotation along a minimal feasible turning radius. Given our grid resolution, we assume a maximal inscribed circle within a cell, $R_{min} = 0.5$. The arc length for a turn of angle $\Delta \theta$ is $s_{arc} = R_{min} |\Delta \theta|$, with a constant curvature $\kappa = 1/R_{min}$. The bending energy penalty for a single junction becomes:
$$E_{turn} = \int_{0}^{s_{arc}} \left( \frac{1}{R_{min}} \right)^2 ds = \frac{|\Delta \theta|}{R_{min}}$$
Thus, the bending energy for the grid-based trajectories is computed as the sum of these discrete turn penalties:
$$E_{bend}^{grid} = \sum_{j} \frac{|\Delta \theta_j|}{R_{min}}$$

\subsection{Empirical Evaluation}

\textbf{Setup.} We evaluated our approach on standard grid maps from the widely used MovingAI benchmark \cite{sturtevant2012benchmarks}: \texttt{Denver} (a city street map), \texttt{arena} (an open space with scattered structural obstacles), and \texttt{empty\_64\_64} (a completely obstacle-free $64 \times 64$ grid). For each map, we utilized the accompanying benchmark scenario files to obtain pairs of start and goal spatial coordinates $(i_0, j_0)$ and $(i_f, j_f)$. The ego-robot is modeled as a circular agent with a radius of $R=1$ (relative to the unit grid cell size).

To ensure a fair geometric baseline comparison between our 3D lattice-based approach and the 2D $2^k$-connected planners, the start and goal heading constraints ($\theta_0$ and $\theta_f$) were relaxed. Since $2^k$ algorithms operate strictly in a spatial domain and allow instantaneous, penalty-free heading changes, forcing rigid terminal orientations on the lattice planner would artificially necessitate large kinematic detours (e.g., looping to align with a final angle). Such detours would disproportionately inflate spatial metrics and skew the comparison. Thus, the search is considered successful once the agent's center reaches the target cell $(i_f, j_f)$ with any valid heading. Strict kinodynamic continuity is naturally maintained by the lattice primitives throughout the rest of the generated trajectory.

Dynamic obstacles are generated procedurally using randomized Depth-First Search (DFS) walks over the state lattice. While these obstacles ignore collisions with the static map and each other, this unstructured generation is a deliberate choice designed to rigorously stress-test the planner. Because our SIPP adaptation relies entirely on the abstraction of safe time intervals per grid cell, the physical coherence or interactions of dynamic hazards are structurally irrelevant to the collision-checking logic. By forcing the algorithm to navigate through densely overlapping, chaotic spatiotemporal constraints, we demonstrate its robustness in a "worst-case" regime, guaranteeing its efficacy in more predictable, physically constrained multi-agent environments. Consistent with standard conventions, all agents disappear from the workspace upon completing their trajectories.


\begin{table}[h!]
\centering
\caption{Merged Benchmark Results for \texttt{Denver\_1\_256} map}
\setlength{\tabcolsep}{8pt}
\renewcommand{\arraystretch}{1.2}
\arrayrulecolor{white}
\setlength{\arrayrulewidth}{2pt}
\begin{tabular}{r | c | c | c | c | c | c}
\toprule
\rowcolor[HTML]{F2F2F2} Obstacles & SIPP-B & SIPP-E & SIPP-2$^2$ & SIPP-2$^3$ & SIPP-2$^4$ & SIPP-2$^5$ \\ \hline
\rowcolor[HTML]{EFEFEF} \multicolumn{7}{l}{\textbf{Path Cost}} \\ \hline
\rowcolor{white} 0 & \cellcolor[HTML]{FFA865}\color{black}0.813x & \cellcolor[HTML]{58E465}\color{black}0.782x & \cellcolor[HTML]{FF3F65}\color{black}1.000x & \cellcolor[HTML]{FF8F65}\color{black}0.826x & \cellcolor[HTML]{DDE465}\color{black}0.791x & \cellcolor[HTML]{3FE465}\color{black}0.782x  \\ \hline
\rowcolor{white} 50 & \cellcolor[HTML]{FF9E65}\color{black}0.821x & \cellcolor[HTML]{64E465}\color{black}0.787x & \cellcolor[HTML]{FF3F65}\color{black}1.000x & \cellcolor[HTML]{FF9065}\color{black}0.829x & \cellcolor[HTML]{DEE465}\color{black}0.795x & \cellcolor[HTML]{3FE465}\color{black}0.786x  \\ \hline
\rowcolor{white} 150 & \cellcolor[HTML]{FF8865}\color{black}0.835x & \cellcolor[HTML]{DEE465}\color{black}0.796x & \cellcolor[HTML]{FF3F65}\color{black}1.000x & \cellcolor[HTML]{FF8F65}\color{black}0.830x & \cellcolor[HTML]{B4E465}\color{black}0.792x & \cellcolor[HTML]{3FE465}\color{black}0.787x  \\ \hline
\rowcolor{white} 320 & \cellcolor[HTML]{FF5D65}\color{black}0.896x & \cellcolor[HTML]{CFE465}\color{black}0.811x & \cellcolor[HTML]{FF3F65}\color{black}1.000x & \cellcolor[HTML]{FF7F65}\color{black}0.854x & \cellcolor[HTML]{EDE465}\color{black}0.814x & \cellcolor[HTML]{3FE465}\color{black}0.803x  \\ \hline
\rowcolor[HTML]{EFEFEF} \multicolumn{7}{l}{\textbf{Trajectory Angularity}} \\ \hline
\rowcolor{white} 0 & \cellcolor[HTML]{E9E465}\color{black}0.123x & \cellcolor[HTML]{B6E465}\color{black}0.100x & \cellcolor[HTML]{FF3F65}\color{black}1.000x & \cellcolor[HTML]{FF6D65}\color{black}0.401x & \cellcolor[HTML]{FFCF65}\color{black}0.155x & \cellcolor[HTML]{3FE465}\color{black}0.076x  \\ \hline
\rowcolor{white} 50 & \cellcolor[HTML]{FFE465}\color{black}0.168x & \cellcolor[HTML]{D2E465}\color{black}0.144x & \cellcolor[HTML]{FF3F65}\color{black}1.000x & \cellcolor[HTML]{FF6465}\color{black}0.479x & \cellcolor[HTML]{FFC165}\color{black}0.201x & \cellcolor[HTML]{3FE465}\color{black}0.110x  \\ \hline
\rowcolor{white} 150 & \cellcolor[HTML]{DDE465}\color{black}0.174x & \cellcolor[HTML]{B5E465}\color{black}0.159x & \cellcolor[HTML]{FF3F65}\color{black}1.000x & \cellcolor[HTML]{FF6365}\color{black}0.505x & \cellcolor[HTML]{FFC365}\color{black}0.223x & \cellcolor[HTML]{3FE465}\color{black}0.137x  \\ \hline
\rowcolor{white} 320 & \cellcolor[HTML]{3FE465}\color{black}0.204x & \cellcolor[HTML]{B1E465}\color{black}0.223x & \cellcolor[HTML]{FF3F65}\color{black}1.000x & \cellcolor[HTML]{FF6365}\color{black}0.542x & \cellcolor[HTML]{FFAB65}\color{black}0.312x & \cellcolor[HTML]{55E465}\color{black}0.205x  \\ \hline
\rowcolor[HTML]{EFEFEF} \multicolumn{7}{l}{\textbf{Angle-Over-Length}} \\ \hline
\rowcolor{white} 0 & \cellcolor[HTML]{CDE465}\color{black}0.102x & \cellcolor[HTML]{8BE465}\color{black}0.080x & \cellcolor[HTML]{FF3F65}\color{black}1.000x & \cellcolor[HTML]{FF7565}\color{black}0.356x & \cellcolor[HTML]{FFD865}\color{black}0.139x & \cellcolor[HTML]{3FE465}\color{black}0.069x  \\ \hline
\rowcolor{white} 50 & \cellcolor[HTML]{E0E465}\color{black}0.140x & \cellcolor[HTML]{A5E465}\color{black}0.117x & \cellcolor[HTML]{FF3F65}\color{black}1.000x & \cellcolor[HTML]{FF6B65}\color{black}0.425x & \cellcolor[HTML]{FFCA65}\color{black}0.181x & \cellcolor[HTML]{3FE465}\color{black}0.099x  \\ \hline
\rowcolor{white} 150 & \cellcolor[HTML]{BBE465}\color{black}0.147x & \cellcolor[HTML]{87E465}\color{black}0.132x & \cellcolor[HTML]{FF3F65}\color{black}1.000x & \cellcolor[HTML]{FF6A65}\color{black}0.448x & \cellcolor[HTML]{FFCF65}\color{black}0.198x & \cellcolor[HTML]{3FE465}\color{black}0.122x  \\ \hline
\rowcolor{white} 320 & \cellcolor[HTML]{3FE465}\color{black}0.170x & \cellcolor[HTML]{B5E465}\color{black}0.190x & \cellcolor[HTML]{FF3F65}\color{black}1.000x & \cellcolor[HTML]{FF6865}\color{black}0.490x & \cellcolor[HTML]{FFB165}\color{black}0.275x & \cellcolor[HTML]{A5E465}\color{black}0.186x  \\ \hline
\rowcolor[HTML]{EFEFEF} \multicolumn{7}{l}{\textbf{Bending Energy}} \\ \hline
\rowcolor{white} 0 & \cellcolor[HTML]{4CE465}\color{black}0.006x & \cellcolor[HTML]{3FE465}\color{black}0.005x & \cellcolor[HTML]{FF3F65}\color{black}1.000x & \cellcolor[HTML]{FF6C65}\color{black}0.356x & \cellcolor[HTML]{FFAC65}\color{black}0.139x & \cellcolor[HTML]{FEE465}\color{black}0.069x  \\ \hline
\rowcolor{white} 50 & \cellcolor[HTML]{4EE465}\color{black}0.008x & \cellcolor[HTML]{3FE465}\color{black}0.007x & \cellcolor[HTML]{FF3F65}\color{black}1.000x & \cellcolor[HTML]{FF6365}\color{black}0.425x & \cellcolor[HTML]{FF9965}\color{black}0.181x & \cellcolor[HTML]{FFC965}\color{black}0.099x  \\ \hline
\rowcolor{white} 150 & \cellcolor[HTML]{3FE465}\color{black}0.009x & \cellcolor[HTML]{43E465}\color{black}0.009x & \cellcolor[HTML]{FF3F65}\color{black}1.000x & \cellcolor[HTML]{FF6065}\color{black}0.448x & \cellcolor[HTML]{FF9365}\color{black}0.198x & \cellcolor[HTML]{FFB965}\color{black}0.122x  \\ \hline
\rowcolor{white} 320 & \cellcolor[HTML]{3FE465}\color{black}0.012x & \cellcolor[HTML]{72E465}\color{black}0.019x & \cellcolor[HTML]{FF3F65}\color{black}1.000x & \cellcolor[HTML]{FF5C65}\color{black}0.490x & \cellcolor[HTML]{FF7E65}\color{black}0.275x & \cellcolor[HTML]{FF9965}\color{black}0.186x  \\ \hline
\bottomrule
\end{tabular}
\label{denver_table}
\end{table}

\begin{figure*}[htbp]
\centering
\setlength{\tabcolsep}{0pt}
\renewcommand{\arraystretch}{1.1}

\begin{tabular}{c}
\rowcolor[HTML]{EFEFEF} \makebox[\textwidth][l]{\hspace{10pt}\textbf{Map: \texttt{Denver\_1\_256}}} \\
\includegraphics[width=0.85\textwidth]{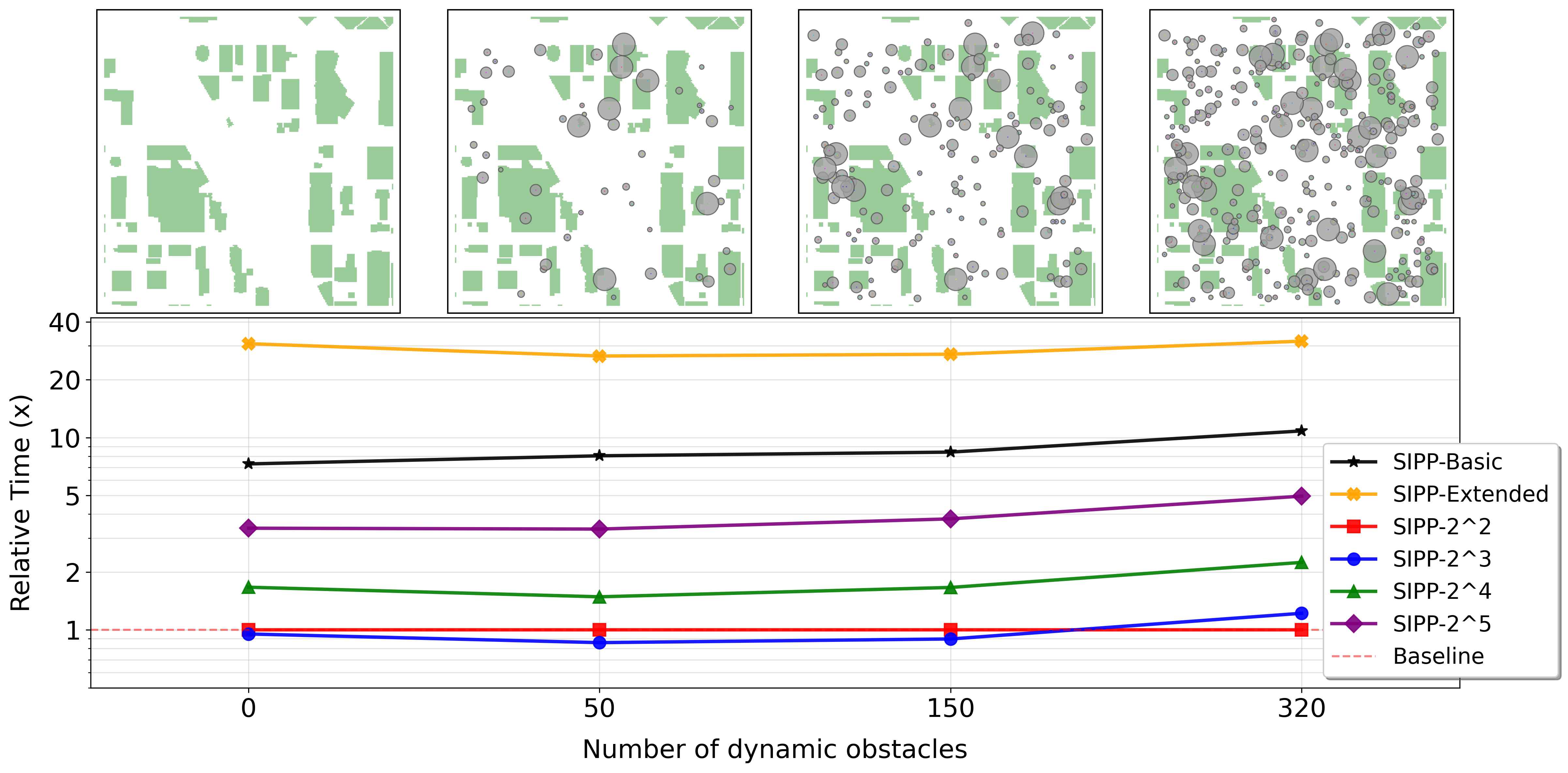} \\

\rowcolor[HTML]{EFEFEF} \makebox[\textwidth][l]{\hspace{10pt}\textbf{Map: \texttt{Empty\_64\_64}}} \\
\includegraphics[width=0.85\textwidth]{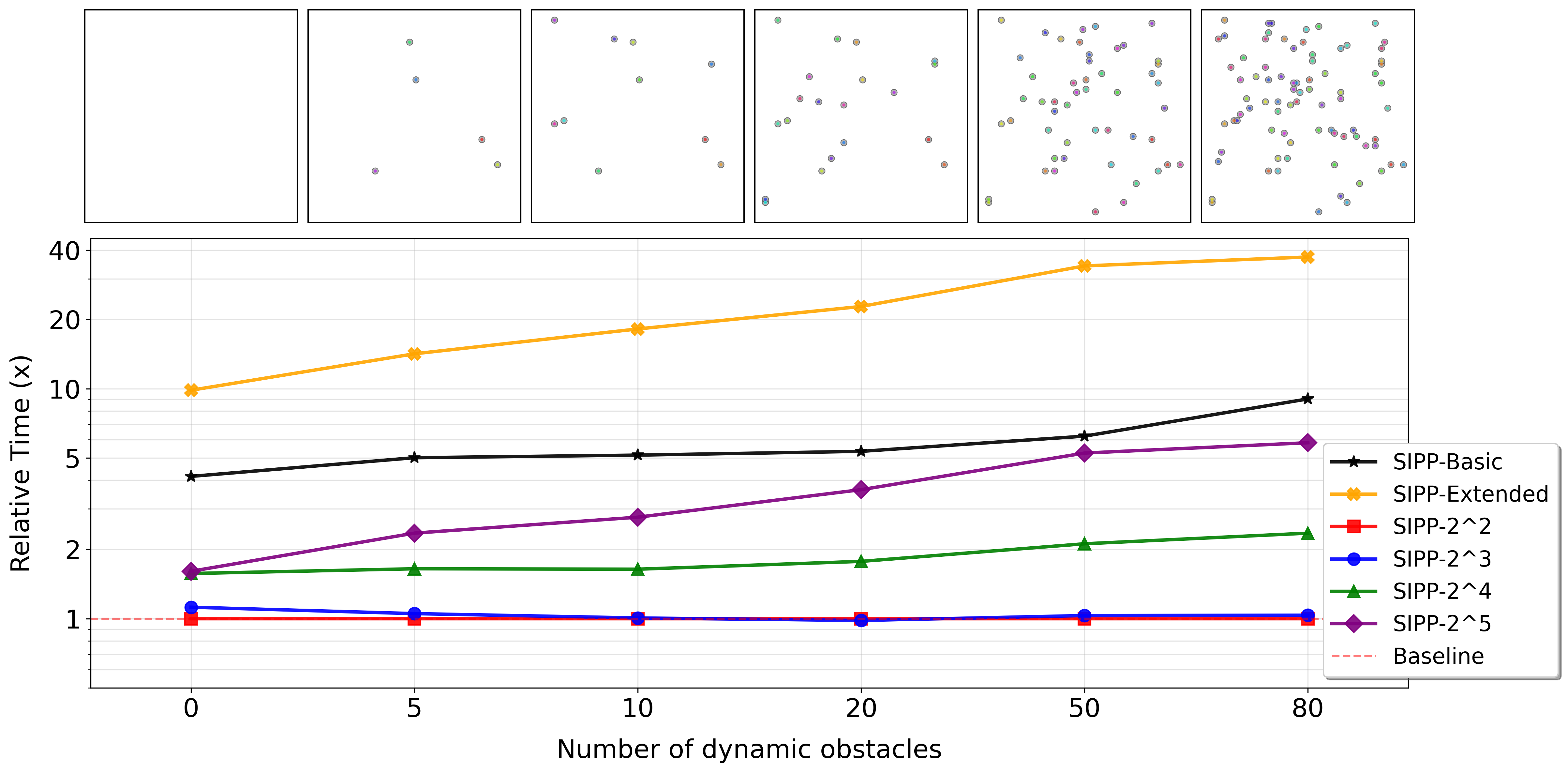} \\

\rowcolor[HTML]{EFEFEF} \makebox[\textwidth][l]{\hspace{10pt}\textbf{Map: \texttt{Arena}}} \\
\includegraphics[width=0.85\textwidth]{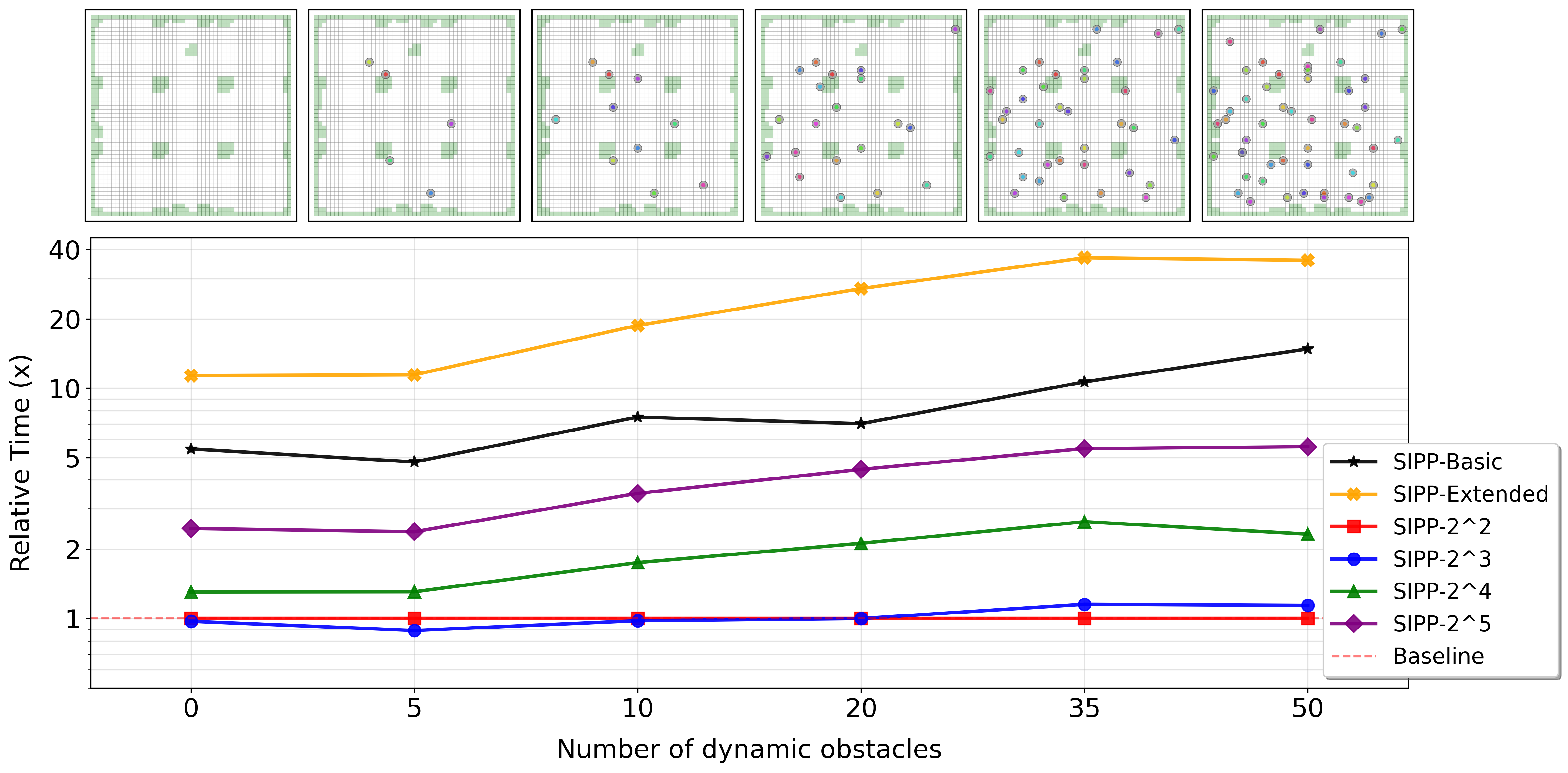} \\
\end{tabular}

\caption{Median runtime comparison across different environments (relative to the runtime of \texttt{SIPP-2\string^2}). The lower -- the better.}
\label{fig:runtime}
\end{figure*}

\begin{figure*}[htbp]
\centering
\setlength{\tabcolsep}{2pt}
\renewcommand{\arraystretch}{1.1}

\begin{tabular}{cc}
\rowcolor[HTML]{EFEFEF} 
\makebox[0.43\textwidth][l]{\hspace{8pt}\textbf{\texttt{SIPP-Basic}}} & 
\makebox[0.43\textwidth][l]{\hspace{8pt}\textbf{\texttt{SIPP-Extended}}} \\
\includegraphics[width=0.43\textwidth]{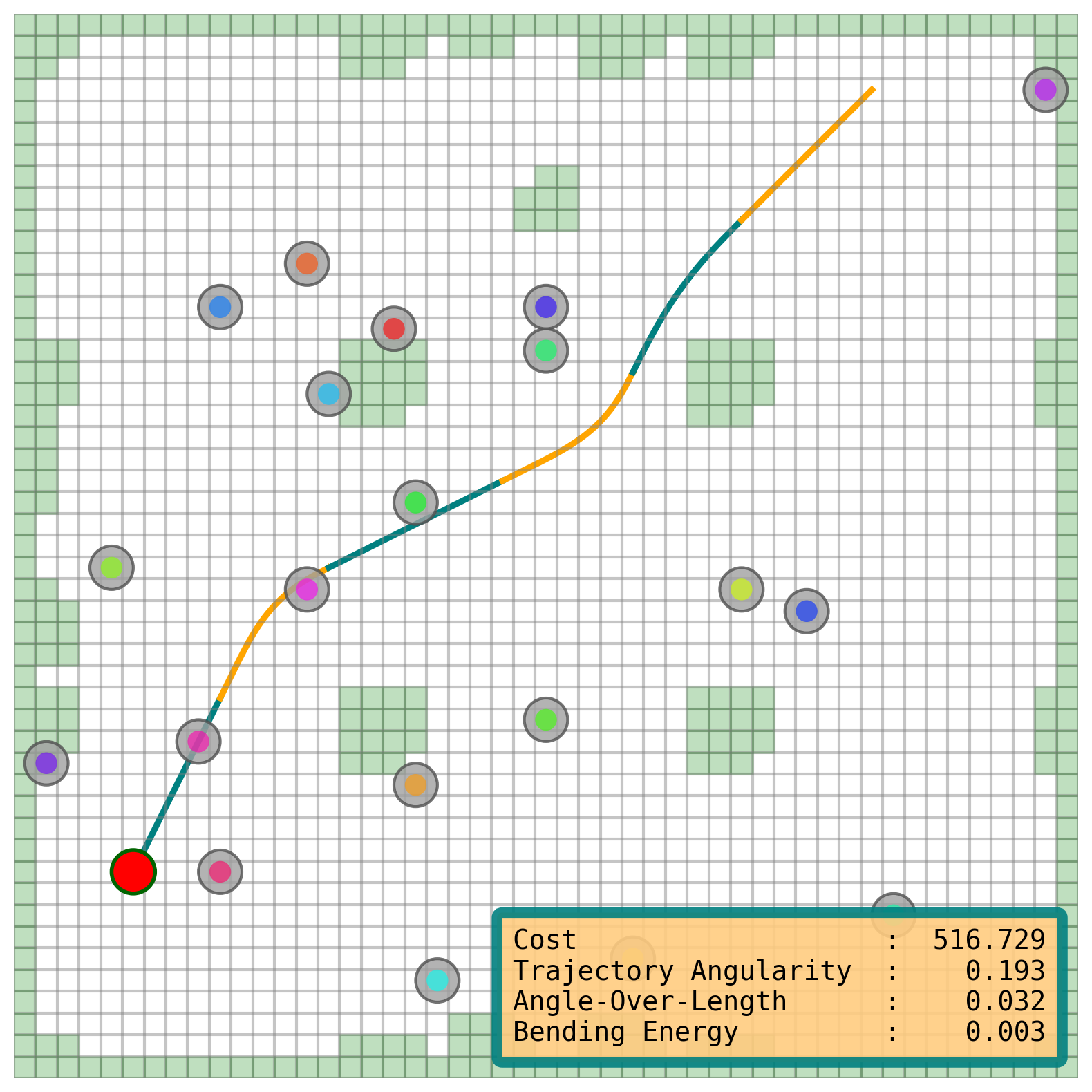} & 
\includegraphics[width=0.43\textwidth]{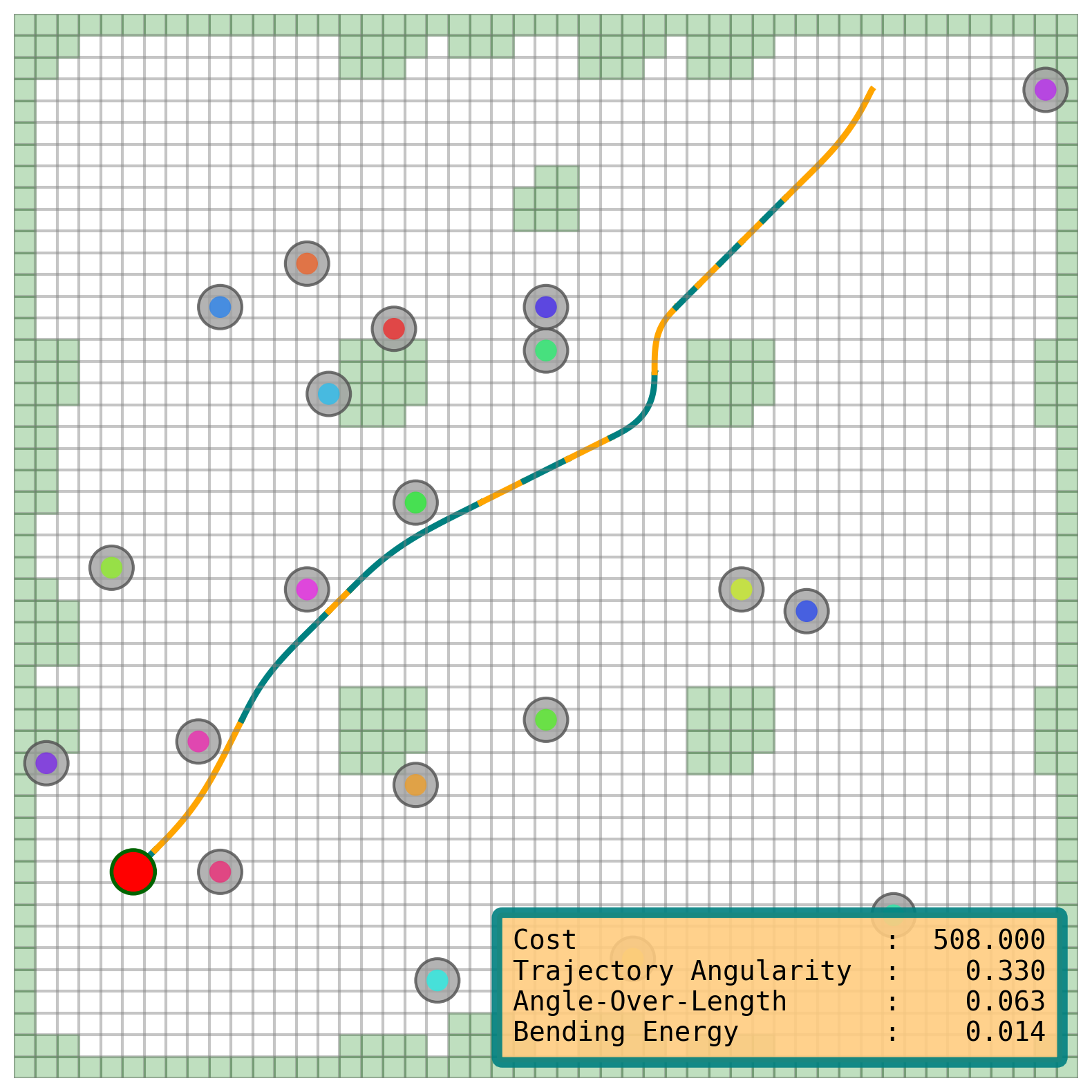} \\[5pt]

\rowcolor[HTML]{EFEFEF} 
\makebox[0.43\textwidth][l]{\hspace{8pt}\textbf{\texttt{SIPP-2\string^2}}} & 
\makebox[0.43\textwidth][l]{\hspace{8pt}\textbf{\texttt{SIPP-2\string^3}}} \\
\includegraphics[width=0.43\textwidth]{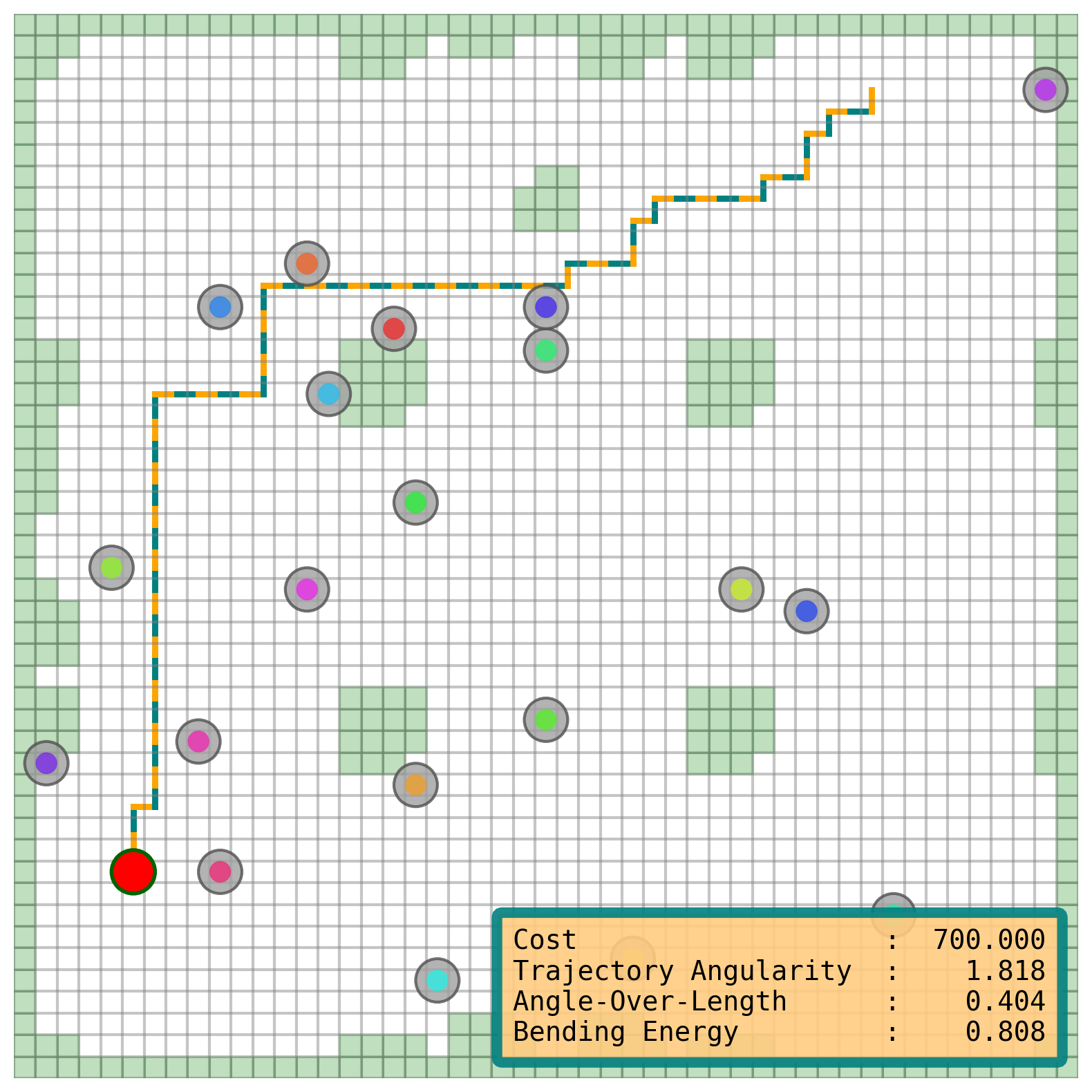} & 
\includegraphics[width=0.43\textwidth]{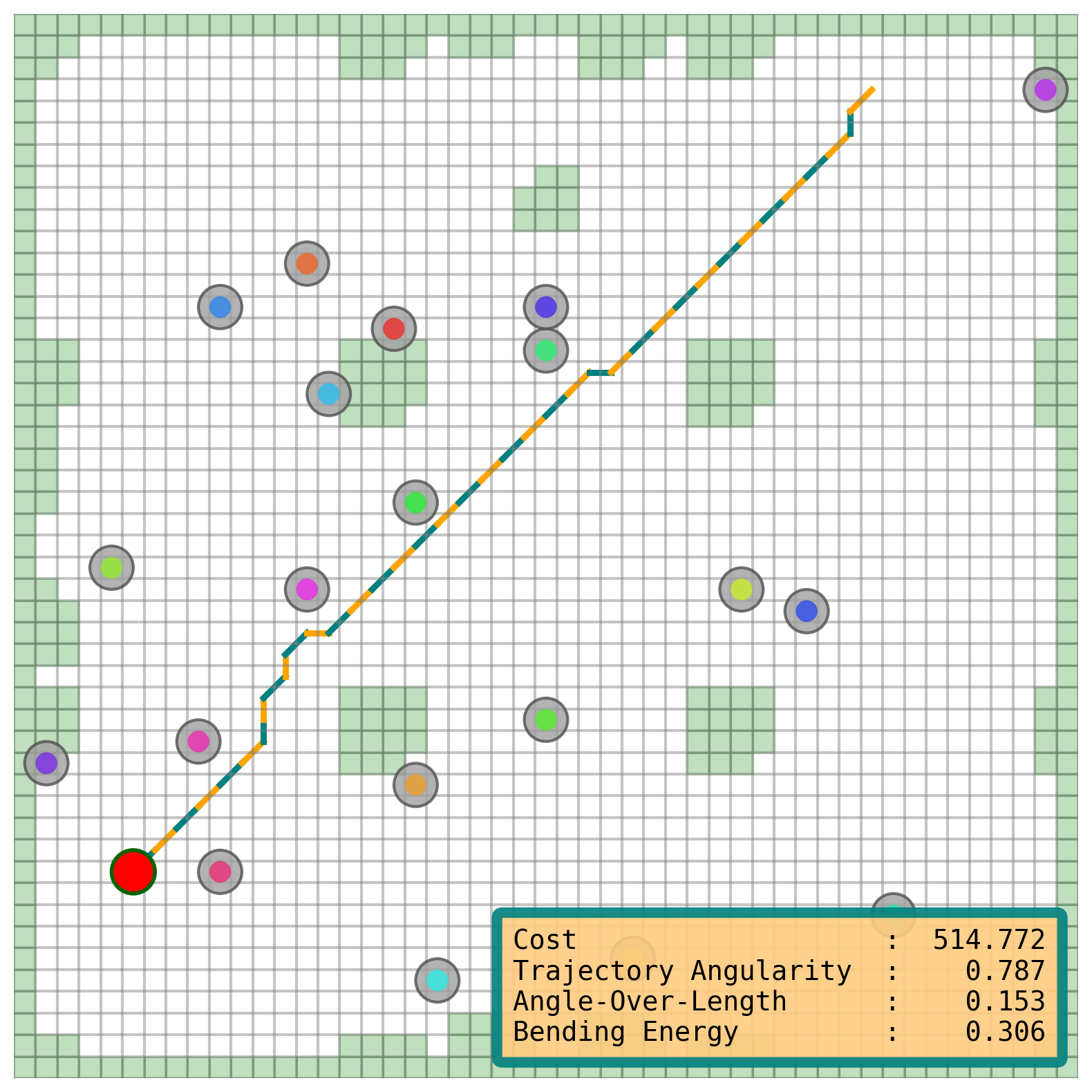} \\[5pt]

\rowcolor[HTML]{EFEFEF} 
\makebox[0.43\textwidth][l]{\hspace{8pt}\textbf{\texttt{SIPP-2\string^4}}} & 
\makebox[0.43\textwidth][l]{\hspace{8pt}\textbf{\texttt{SIPP-2\string^5}}} \\
\includegraphics[width=0.43\textwidth]{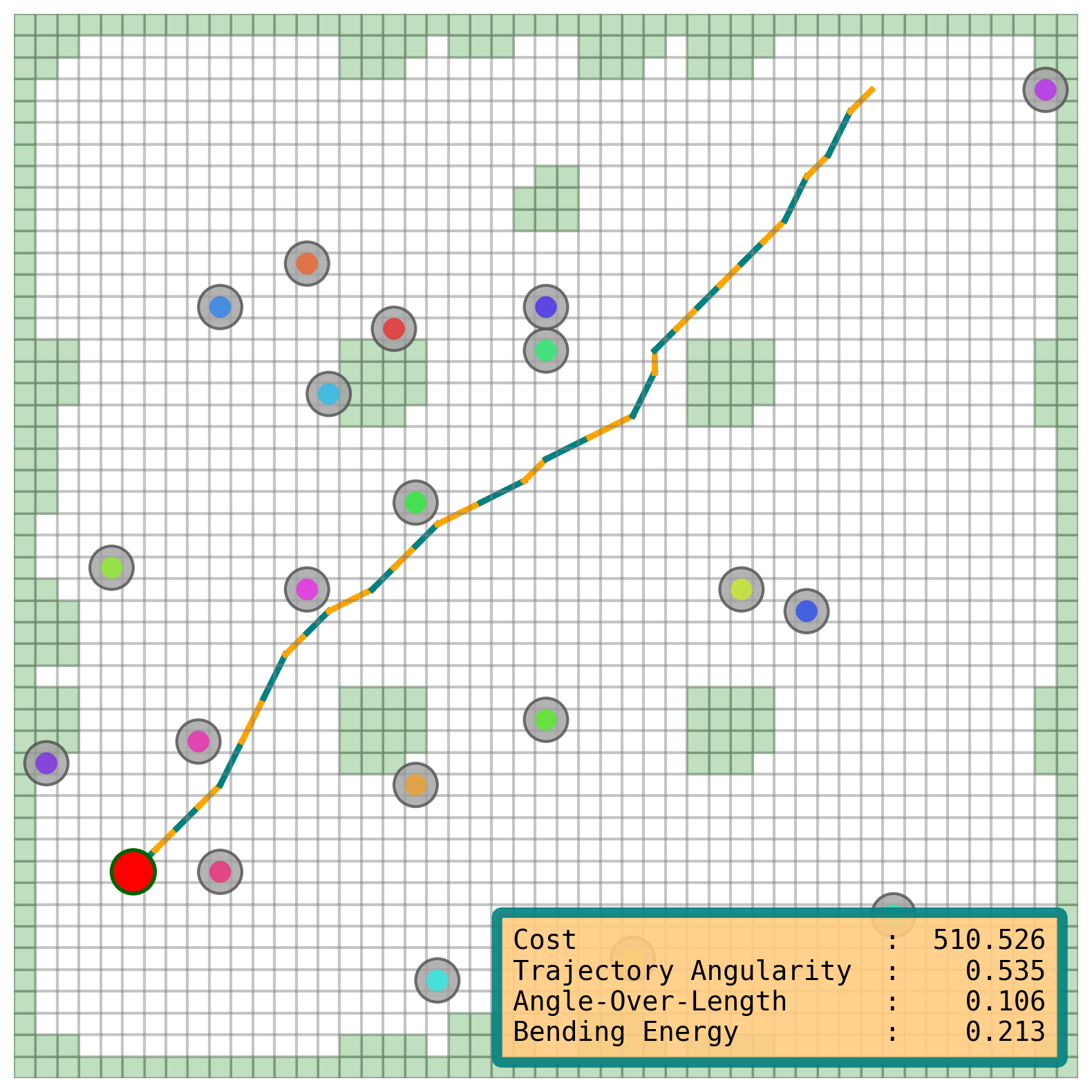} & 
\includegraphics[width=0.43\textwidth]{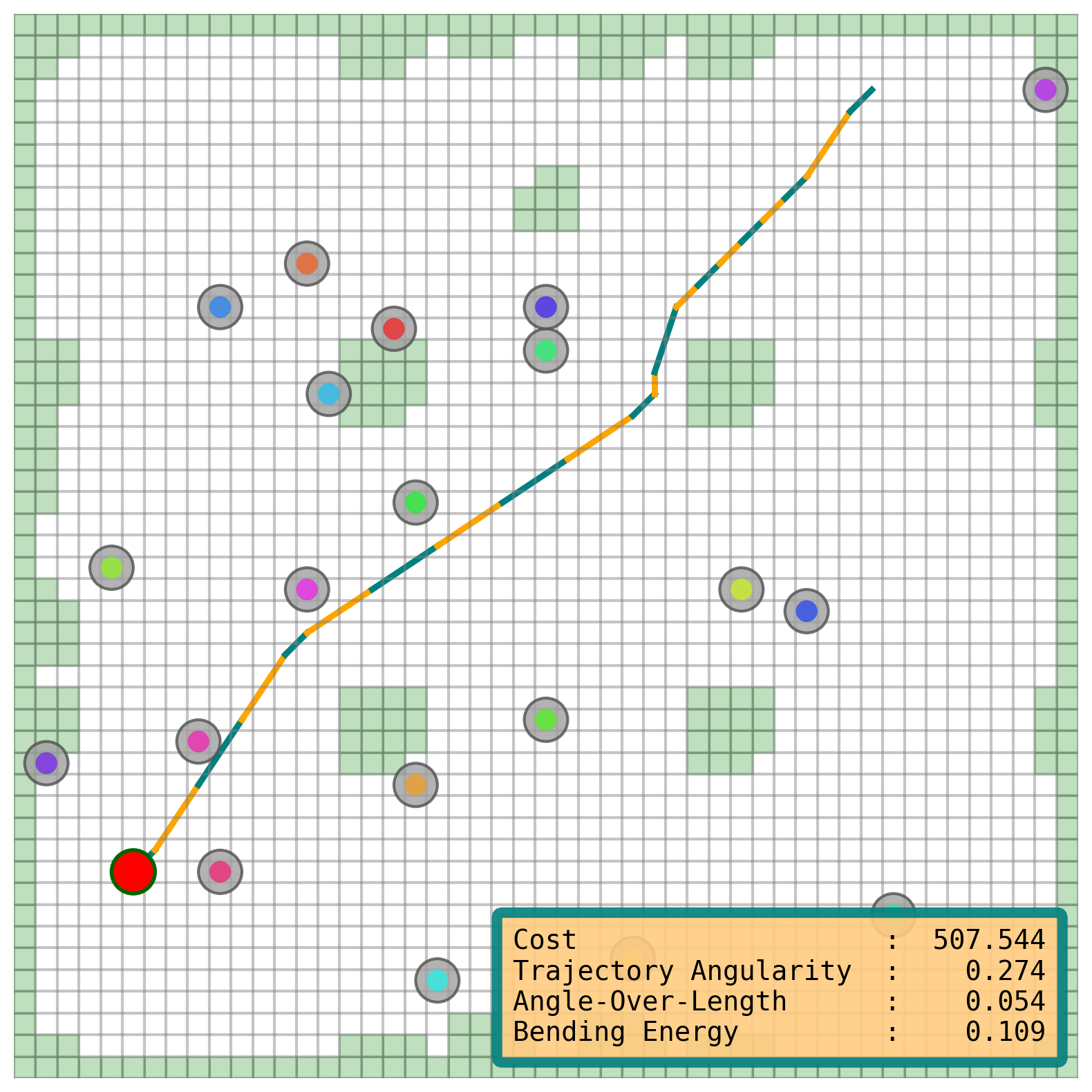} \\
\end{tabular}

\caption{Paths for the \texttt{Arena} map with 20 dynamic obstacles of radius $1$ cell. Planning is performed between cells $(39, 5)$ and $(3, 39)$; the snapshot is shown at $t = t_0 = 0$.}
\label{fig:search_examples}
\end{figure*}

\noindent
\textbf{Results.} Metric values for each planning instance were normalized relative to the standard 4-connected search (\texttt{Grid-2\string^2}) baseline, with aggregated results representing the medians of these normalized values. Detailed geometric and cost metrics for the \texttt{Denver\_1\_256} map are presented in Table~\ref{denver_table}. Computational performance across all environments is summarized in Figure~\ref{fig:runtime}, and qualitative trajectory examples are depicted in Figure~\ref{fig:search_examples}. Since SIPP is provably optimal, each generated path is strictly optimal with respect to its specific control set.

\textbf{Search Time.} As expected, lattice-based algorithms incur higher computational costs than \texttt{Grid-2\string^k} variants (Fig.~\ref{fig:runtime}). This overhead stems naturally from expanding nodes in a 3D state space $(i, j, \theta)$ compared to the 2D spatial domain $(i, j)$, representing a direct and necessary trade-off between search speed and kinodynamic smoothness.

\textbf{Trajectory Quality.} The primary advantage of the lattice-based approach lies in its physical executability. As shown in Table~\ref{denver_table}, lattice algorithms demonstrate a massive reduction in Bending Energy ($E_{bend}$) compared to all \texttt{Grid-2\string^k} variants. While $2^k$-connected grids require robots to perform severe, non-smooth point-turns, lattice primitives replace these junctions with continuous curves, yielding paths that are inherently stable and ready for low-level control systems.

\FloatBarrier

\FloatBarrier

\section{Conclusion}

In this paper, we presented a dynamic path planning approach integrating state lattices with the SIPP algorithm. By rasterizing the spatial-temporal swept volumes of moving obstacles directly onto the grid, we enabled efficient, sampling-free collision checking.

Our evaluation reveals that while highly connected grids (e.g., \texttt{Grid-2\string^5}) perform exceptionally well on spatial metrics due to their dense branching, they produce kinematically unfeasible, jagged paths. Nevertheless, they serve as a geometric benchmark: by progressively enriching the lattice control set, we can successfully match these superior routing metrics while guaranteeing physically executable, smooth trajectories.

However, the increased dimensionality of the state lattice inevitably impacts search time. To address this, future work will explore adapting the MeshA* algorithm \cite{mesha_star} to evaluate bundles of primitives simultaneously, aiming to drastically reduce the branching factor while preserving optimality. Moving forward, we plan to validate our framework via closed-loop physical deployments against planners such as TEB or MPC, and extend this approach to Multi-Agent Path Finding (MAPF).

\bibliography{sn-bibliography}

\end{document}